\documentclass[letterpaper, 10 pt, conference]{ieeeconf}  %

\IEEEoverridecommandlockouts                              %
                                                          
\usepackage{bm}
\usepackage{cite}
\usepackage{amsmath,amssymb,amsfonts,mathcomp}
\usepackage{algorithmic}
\usepackage{graphicx}
\usepackage{float}
\usepackage{here}
\usepackage{textcomp}
\usepackage{color}
\usepackage{algorithmic}
\usepackage{algorithm}
\usepackage{booktabs}
\usepackage{comment}
\usepackage{url}
\usepackage{hyperref}

\begin{document}

\title{D-SLAMSpoof: An Environment-Agnostic LiDAR Spoofing Attack using Dynamic Point Cloud Injection}

\author{Rokuto Nagata$^{1}$, Kenji Koide$^{2}$, Kazuma Ikeda$^{1}$, Ozora Sako$^{1}$, and Kentaro Yoshioka$^{1}$% <-this % stops a space
%\thanks{*This work was not supported by any organization}% <-this % stops a space
\thanks{$^{1}$ Department of Electronics and Electrical Engineering, Keio University}
\thanks{$^{2}$ Department of Information Technology
and Human Factors, the National Institute of Advanced Industrial Science and Technology}
}

\maketitle

\begin{abstract}

In this work, we introduce Dynamic SLAMSpoof (D-SLAMSpoof), a novel attack that compromises LiDAR SLAM even in feature-rich environments. The attack leverages LiDAR spoofing, which injects spurious measurements into LiDAR scans through external laser interference. By designing both spatial injection shapes and temporally coordinated dynamic injection patterns guided by scan-matching principles, D-SLAMSpoof significantly improves attack success rates in real-world, feature-rich environments such as urban areas and indoor spaces, where conventional LiDAR spoofing methods often fail.
Furthermore, we propose a practical defense method, ISD-SLAM, that relies solely on inertial dead reckoning signals commonly available in autonomous systems. We demonstrate that ISD-SLAM accurately detects LiDAR spoofing attacks, including D-SLAMSpoof, and effectively mitigates the resulting position drift.
Our findings expose inherent vulnerabilities in LiDAR-based SLAM and introduce the first practical defense against LiDAR-based SLAM spoofing using only standard onboard sensors, providing critical insights for improving the security and reliability of autonomous systems.

\end{abstract}

\section{Introduction}
\label{introduction}
\vspace{-0.05in}

Autonomous navigation enables robots to operate effectively across diverse tasks and environments, from delivery systems to fully autonomous vehicles, thereby greatly expanding their range of applications and accelerating the advancement of automation~\cite{rubio2019review}. Among the sensing modalities that support such autonomy, LiDAR (Light Detection And Ranging) has emerged as a central technology. Since the 2007 DARPA Urban Challenge, where LiDAR-based navigation demonstrated high reliability~\cite{urmson2007tartan}, LiDAR has become the de facto standard for systems requiring robust self-localization, including modern self-driving vehicles~\cite{permitted_ad}.

\begin{figure}[t]
  \centering
  \includegraphics[width=\linewidth]{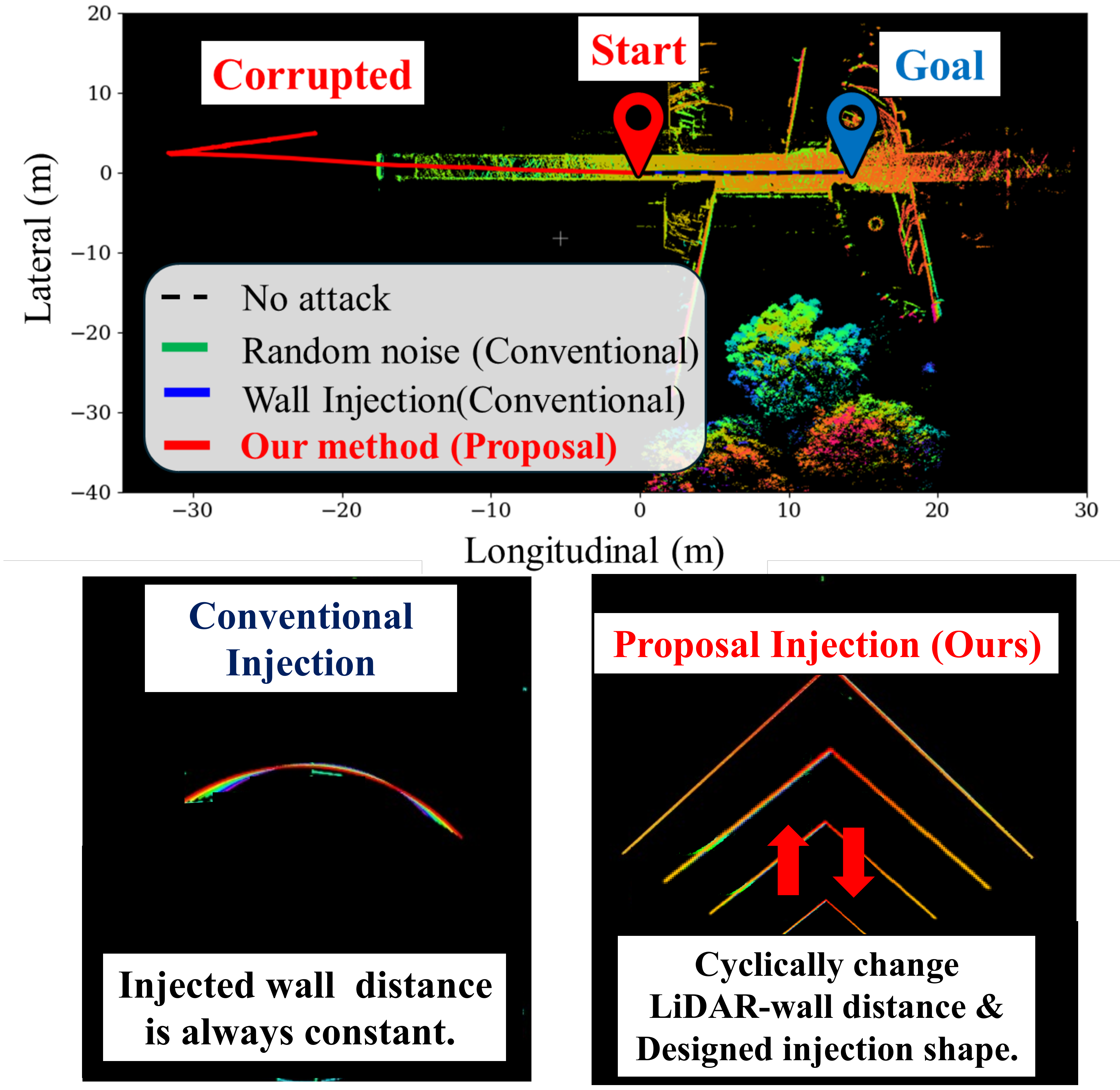}
  \vspace{-0.3in}
  \caption{(Top) Real-world spoofing attack demonstration under a feature-rich environment is shown. Only D-SLAMSpoof (red line) successfully disrupted LiDAR odometry estimated by FAST-LIO2~\cite{fastlio2}, whereas conventional attack methods (blue and green lines) had no effect on the estimation. (Bottom) While conventional point cloud injection injects a cylindrical wall at a fixed distance, our method strengthens the attack by designing efficient point cloud shapes and periodically varying their distance.}
  \label{header_realworld_experiment}
\end{figure}

While LiDAR is a core technology supporting the safe navigation of autonomous robots, it is also vulnerable to a critical threat known as LiDAR spoofing, in which external laser emissions inject falsified range measurements into the sensor~\cite{petit2015remote, cao2019adversarial, sato2024lidar}. uch manipulation of the point cloud can severely degrade self-localization, potentially leading to hazardous behaviors such as unintended trajectory deviations or collisions. Consequently, this vulnerability poses a fundamental threat to the safety and reliability of autonomous systems that depend on LiDAR-based localization.

Although the potential risk of localization attacks via LiDAR spoofing has been demonstrated, prior research has been restricted to environments with sparse geometric features. This is because scan matching is highly reliable in geometrically rich environments, such as urban areas and indoor spaces, making LiDAR-based localization difficult to disrupt. As a result, the vulnerabilities of LiDAR-based SLAM in geometrically rich real-world environments remain insufficiently understood. Fukunaga et al.~\cite{fukunagarandom} showed that injecting random noise into LiDAR measurements can corrupt SLAM-generated maps and degrade localization accuracy. However, this method relies on limited opportunities during map construction, and the induced self-localization error is primarily confined to the vertical (z) axis, limiting its threat. Furthermore, while Nagata et al. and Tanaka et al.~\cite{ nagata_slamspoof_journal, tanaka2025wip} demonstrated real-time attacks, their success still depended on selecting environments with few structural features, leaving the broader security risks of LiDAR-based SLAM in realistic, feature-rich settings largely unexplored.

On the defensive side, Nagata et al.~\cite{nagata_slamspoof_journal} proposed DefenSLAM, which detects and removes injected point clouds, reporting high error suppression. However, this method requires access to the LiDAR's raw waveform data, making it inapplicable to most commercial off-the-shelf systems.

In summary, existing attack research has been biased toward environments with limited geometric features, while defensive methods have lacked practicality. This highlights the need to systematically investigate the vulnerabilities of LiDAR SLAM in geometrically rich, real-world environments and to establish a defensive method that can be implemented to standard sensors commonly available on autonomous platforms.

In this work, we aim to provide a comprehensive analysis of the security risks inherent in LiDAR SLAM by proposing a novel attack method, Dynamic SLAMSpoof (D-SLAMSpoof), which is designed to be independent of environmental geometry. This method disrupts LiDAR-based localization with significantly higher success rates than conventional approaches, as demonstrated on open datasets and real-world experiments. Our results reveal that LiDAR SLAM is intrinsically vulnerable to point cloud manipulation even under conditions previously considered robust and beyond the scope of existing attack models. Furthermore, as part of our commitment to responsible research, we propose a defense system, ISD-SLAM, which is both practical and operates using only standard onboard sensors such as an IMU and inertial odometry. We show that ISD-SLAM detects spoofing-induced localization errors with over 90\% accuracy and effectively suppresses abrupt localization drift. The main contributions of this study are as follows.

\noindent\textbf{Environment-Agnostic Attack:} 
    We propose D-SLAMSpoof, a LiDAR spoofing attack that jointly optimizes the geometry and temporal pattern of the injected point cloud to mislead the scan matching process. The proposed method significantly improves attack success rates across multiple SLAM algorithms, particularly in geometrically rich environments where existing approaches are less effective.
\noindent\textbf{Evaluation in Realistic Environments:} 
    We evaluate D-SLAMSpoof through both simulation and real-world experiments on four representative LiDAR SLAM systems, including state-of-the-art implementations. The results show consistently stronger attack performance than existing spoofing methods, even in environments with abundant geometric features.
\noindent\textbf{Deployable Defense Without Additional Hardware:} 
    We propose ISD-SLAM, a defense method that combines inertial anomaly detection with adaptive switching to inertial navigation. To the best of our knowledge, this is the first defense against LiDAR-based SLAM spoofing that goes beyond detection to actively mitigate localization drift using standard onboard sensors, without requiring specialized hardware or access to raw LiDAR waveforms.

\section{Related work and Background}
\label{related_work}

\begin{figure*}[tb]
  \centering
  \includegraphics[width=\linewidth]{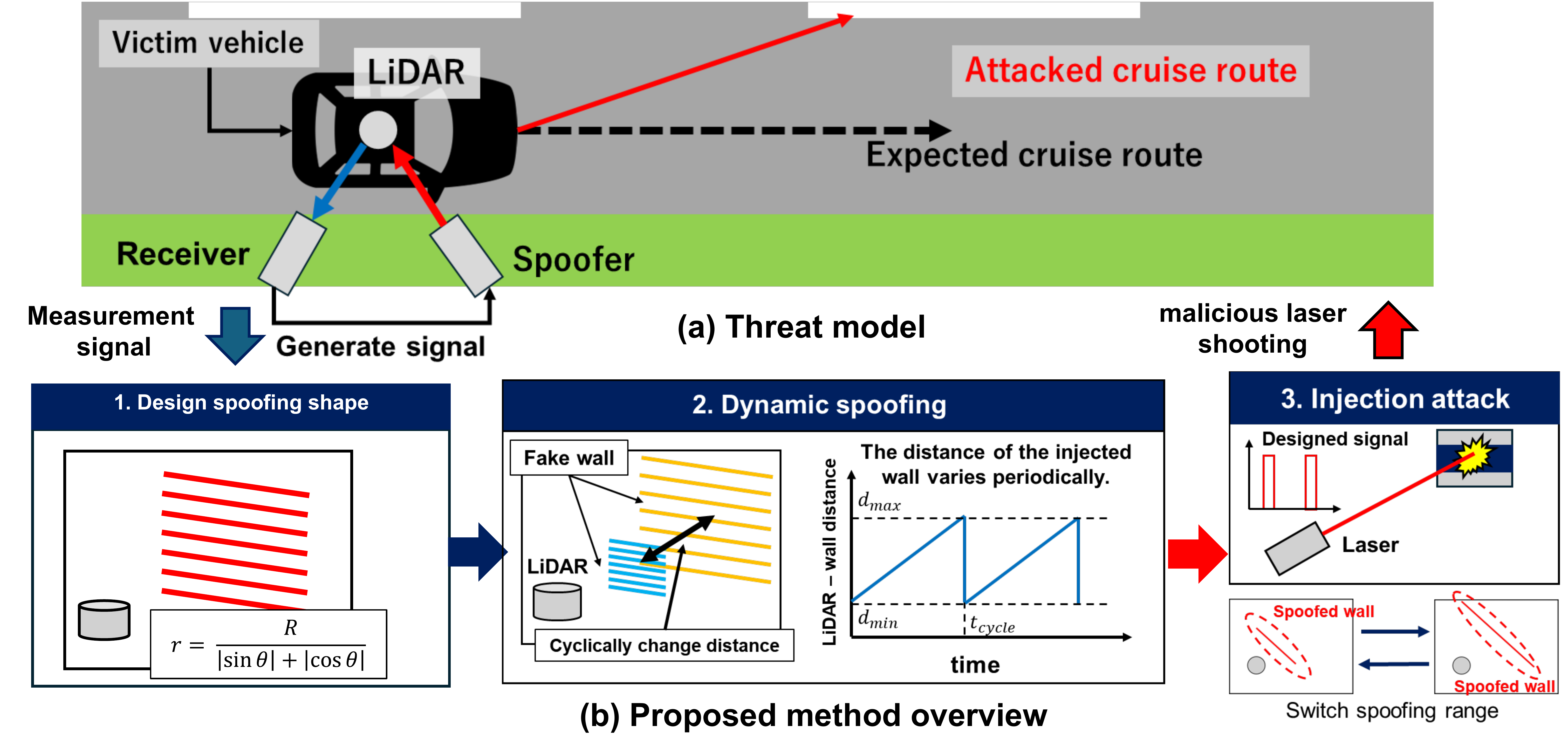}
  \vspace{-0.35in}
  \caption{An overview of our proposed method. (a) Threat model overview. The attacker aims to induce errors in the self-localization of the target vehicle by placing a roadside laser source that tampers the point cloud through malicious laser shooting. The attacker is assumed to know the target's route in advance. (b) Overview of D-SLAMSpoof. The attacker designs the injected point cloud shape based on geometric constraints in scan matching and periodically moves the point cloud in the radial direction to maximize inter-frame displacement, thereby strengthening the attack. Details for designing the injection shape is detailed in \S~\ref{propose_injection_shape}, and the method for moving the injected point cloud is described in \S~\ref{propose_injection_dynamic}.}
  \vspace{-0.2in}
  \label{fig_threat_model}
\end{figure*}

\subsection{LiDAR-based localization}
\label{LiDAR_based_localization}

LiDAR-IMU SLAM techniques~\cite{fastlio2, koide2024glim} are widely adopted, fusing LiDAR and IMU measurements to enhance localization robustness. Such tightly coupled systems are generally regarded as resilient to environmental disturbances and sensor noise. However, their robustness against adversarial point cloud manipulation has not been thoroughly examined. Recent studies suggest that even widely used and state-of-the-art SLAM algorithms may remain vulnerable under carefully crafted spoofing conditions.

\subsection{LiDAR spoofing attacks to localization}
\label{prior_attempts_localization_attack}

LiDAR spoofing attacks falsify measurements by emitting malicious laser pulses and are broadly categorized into injection attacks (inserting false point clouds)~\cite{petit2015remote, shin2017illusion, cao2019adversarial} and removal attacks (suppressing legitimate returns)~\cite{shin2017illusion, cao2023you, sato2024lidar}. Recent research~\cite{sato2024lidar} has shown that nearly all point clouds within an 80-degree field of view can be replaced.
Several studies have demonstrated that LiDAR spoofing can degrade SLAM accuracy~\cite{yoshida2022adversarial, fukunagarandom, nagata_slamspoof_journal, tanaka2025wip}. However, early work was limited to laboratory settings or 2D LiDAR systems~\cite{yoshida2022adversarial}. Subsequent research showed that random point injection can contaminate map construction~\cite{fukunagarandom}, and more recent studies demonstrated real-time position manipulation~\cite{nagata_slamspoof_journal, tanaka2025wip, zhang2025prepared}.
Nevertheless, the effectiveness of these attacks often depends on environmental geometry. Prior analyses indicate that ICP stability is highly influenced by geometric constraints and structural diversity~\cite{laconte2023toward, zhang2025prepared}. Consequently, successful real-world attacks have largely been confined to geometrically sparse environments, leaving the security of LiDAR-based SLAM under structurally rich conditions insufficiently explored. In this work, we demonstrate that LiDAR spoofing can induce significant localization errors even in urban and indoor environments, which are typically considered robust due to their abundant geometric features.

\subsection{LiDAR spoofing defenses}\label{detection method}

Several defensive methods against LiDAR spoofing have been proposed; however, most focus on protecting object detection rather than localization or SLAM. For instance, methods like CALRO, Adopt, and TC2~\cite{sun2020towards, cho2023adopt, you2021temporal} are designed to defend perception pipelines and are not intended to safeguard scan-matching-based self-localization.
Suzuki et al.~\cite{suzuki2025_from} proposed detecting spoofing by analyzing raw LiDAR waveform signals. While effective, this approach requires access to low-level sensor data that is typically unavailable in commercial LiDAR systems, limiting its practical deployability. Furthermore, anomaly monitoring strategies~\cite{arana2019efficient, alsayed2017failure} can detect abnormal system behavior but do not provide mechanisms for correcting erroneous state estimates or actively protecting the estimated trajectory.
Consequently, there remains no defense specifically designed to protect LiDAR-based SLAM from LiDAR spoofing attacks that is deployable using only standard commercial sensor configurations. To address this gap, we propose ISD-SLAM, the first practical defense against LiDAR-based SLAM spoofing that is fully deployable on commercial systems without requiring additional hardware or access to raw LiDAR waveforms.

\subsection{Threat model}
\label{threat_model}
In this study, we assume an attacker capable of placing a single spoofing device along the target vehicle’s route (Fig.~\ref{fig_threat_model}(a)). The target platform is equipped with a 3D LiDAR and an IMU. The attacker is assumed to know the vehicle’s route and LiDAR specifications but has no access to internal software or onboard sensor data.
The attacker’s objective is to disrupt self-localization by injecting falsified point cloud measurements from an external source, thereby causing the vehicle to deviate from its intended trajectory. Prior analysis by Sato et al.~\cite{sato2021dirty} indicates that localization errors as small as 0.29,m on general roads or 0.74,m on highways may lead to lane departure, posing a direct risk of collision.

\section{Methodology: D-SLAMSpoof Framework}
\label{methodology}

\subsection{D-SLAMSpoof Overview}
\label{attack design policy}

D-SLAMSpoof is a dynamic LiDAR spoofing framework that targets the optimization mechanism of scan-matching-based SLAM, even in geometrically rich environments. Unlike conventional spoofing methods that rely on static or randomly injected point clouds, D-SLAMSpoof constructs temporally coordinated adversarial point clouds that systematically bias pose estimation during scan matching. By jointly manipulating spatial geometry and temporal consistency, the attack induces persistent pose estimation errors independent of environmental feature density.

The goal of this work is to rigorously evaluate the threat posed by such coordinated spoofing in environments traditionally considered robust for LiDAR-based SLAM. To this end, D-SLAMSpoof integrates two complementary strategies: (1) forging geometric constraints that skew the optimization landscape, and (2) dynamically amplifying inter-frame miscorrespondences to accumulate estimation drift.

\begin{itemize}
\item \textbf{Constraint-Forging Injection:} LiDAR SLAM estimates sensor pose by minimizing geometric residuals between consecutive point clouds. Exploiting this formulation, we design adversarial point cloud structures that introduce strong pose estimation constraints in a specific direction, overwhelming constraints of correct point correspondences.
\item \textbf{Oscillating Injection:} Large pose errors often arise when incorrect correspondences accumulate across frames. However, static injection creates only small inter-frame differences and struggles to cause persistent estimation errors. We therefore periodically oscillate the injected point cloud in the radial direction, inducing controlled inter-frame discrepancies. This motion destabilizes scan matching convergence and enables persistent trajectory deviation.
\end{itemize}

Fig.~\ref{fig_threat_model}(b) shows an overview of this methodology, and the following sections will detail these two strategies.

\subsection{Constraint-Forging Injection}
\label{propose_injection_shape}

Following the strategy outlined above, we design the injected point cloud to exploit the directional consistency of geometric constraints in scan-matching-based SLAM.
Prior injection attacks~\cite{nagata_slamspoof_journal} did not explicitly consider the geometric structure of injected points and instead employed simple cylindrical patterns. Consequently, the induced false constraints were distributed across multiple directions, resulting in small and irregular pose perturbations that were largely absorbed by the inherent robustness of SLAM.
To overcome this limitation, we propose Constrain-Forging Injection, which injects a point cloud designed to align the directions of false geometric constraints. 

Specifically, we define the injected shape using the following polar equation:
\begin{align}
    d_{\text{fake}} = \frac{S}{|\sin{\theta'}|+|\cos{\theta'}|}
\end{align}
Here, $S$ is a scaling constant. This equation represents the boundary of a square centered at the origin in polar coordinates. By rotating the polar angle, it can selectively generate fake planar or corner-like structures. Specifically, in its base orientation, an L-shaped wall bent at a right angle and facing the spoofer is injected from the LiDAR's perspective. Alternatively, by rotating the polar equation by $\frac{\pi}{4}$ radians, a planar injection shape is realized where the square's edge aligns perpendicularly to the LiDAR's ranging direction.

Because the generated structure concentrates residuals along specific orientations, scan matching repeatedly favors incorrect correspondences in a consistent direction. This accumulated directional bias induces persistent pose drift, eventually exceeding the correction capability of SLAM’s robustness mechanisms and leading to localization failure.

\subsection{Oscillating Injection}
\label{propose_injection_dynamic}
To effectively destabilize scan-matching-based localization, we introduce Oscillating Injection, which periodically moves the injected point cloud along the radial direction. Static injection produces limited inter-frame variation and therefore induces only small, transient pose perturbations. In contrast, controlled oscillatory motion maximizes inter-frame geometric discrepancies, increasing the likelihood of persistent miscorrespondences and accumulated pose drift.

To realize effective oscillating injection, we must carefully design the movement pattern, as too large point cloud movement can be detected and rejected by SLAM's outlier filtering mechanisms. Therefore, the movement must satisfy the following two constraints to maximize attack effectiveness while evading detection.

\noindent\textbf{1. Constraint of LiDAR's Ranging Capability:}
Since the fake point cloud will be ignored if it exceeds the sensor's effective ranging distance, the injection distance must always remain within this range. Therefore, the maximum point cloud injection distance $d_{\text{max}}$is determined by the LiDAR's maximum measurement distance $M_{\text{measure}}$ to satisfy the following condition:
\vspace{-0.1in}
\begin{align}
d_\text{max} \leqq M_{\text{measure}}
\end{align}

\noindent \textbf{2. Constraint of Maximum Correspondence Distance:} SLAM algorithms implement outlier rejection based on a maximum correspondence distance threshold. In theory, moving the injected point cloud significantly between frames should create correspondences with large geometric errors, inducing a severe pose estimation error. However, if the inter-frame movement of the injected point cloud exceeds the maximum correspondence distance, it will be rejected from the correspondence set, failing to induce a miscorrespondence. Therefore, it is subject to the constraint expressed by the following equation:
\begin{align}
\frac{d_{max}-d_{min}}{t_{cycle}}\Delta t \leqq M_{\text{corr}}
\end{align}
Here, $d_\text{min}$, $M_{\text{corr}}$ are the minimum point cloud injection distance and max correspondence distance, $\Delta t$ is the interval at which the LiDAR outputs a point cloud (typically 0.05 to 0.1 seconds), and $t_{\text{cycle}}$ is the period for moving the point cloud. 
Under the constraint from the maximum correspondence distance $M_{\text{corr}}$, the period$t_{\text{cycle}}$ that maximizes the attack capability is determined by the following equation.
\vspace{-0.1in}
\begin{align}
t_{cycle} = \frac{(d_{max}-d_{min})}{M_\text{corr}} \Delta t
\end{align}

To satisfy these two conditions, we design the maximum injection distance, the minimum injection distance, and the period $t_{\text{cycle}}$. Ultimately, by linearly varying the distance of the fake point cloud, $d_{\text{fake}}$, over time, we realize the Oscillating Injection with maximum attack effectiveness:
\begin{align}
    d_\text{fake} = \frac{d_{max}-d_{min}}{t_{cycle}}t +d_{\text{min}} (0 \leqq t \leqq t_{\text{cycle}})
\end{align}

\section{Methodology: Practical defense framework}
\label{methodology_defence}
Despite prior efforts, practical and deployable defenses for protecting LiDAR-based SLAM from spoofing remain limited. We propose ISD-SLAM (Inertial Switching Defense for SLAM), a defense framework that detects spoofing-induced localization errors and dynamically switches to inertial dead-reckoning when an attack is detected.
\vspace{-0.05in}
\subsection{Detection of Spoofing-Induced Localization Errors}
\label{propose_inertial_detection}
ISD-SLAM detects spoofing by monitoring inconsistencies between SLAM-based pose estimates and inertial dead-reckoning. Inertial sensors, such as IMUs and wheel odometry, estimate ego-motion independently of environmental observations and therefore remain unaffected by point cloud manipulation.

Spoofing-induced localization errors typically manifest as (1) orientation inconsistencies and (2) abnormal translational velocity deviations. Accordingly, our detection framework consists of two steps: estimating these error components and integrating them into a unified detection metric.

\begin{figure*}[tb]
  \centering
  \includegraphics[width=0.95\linewidth]{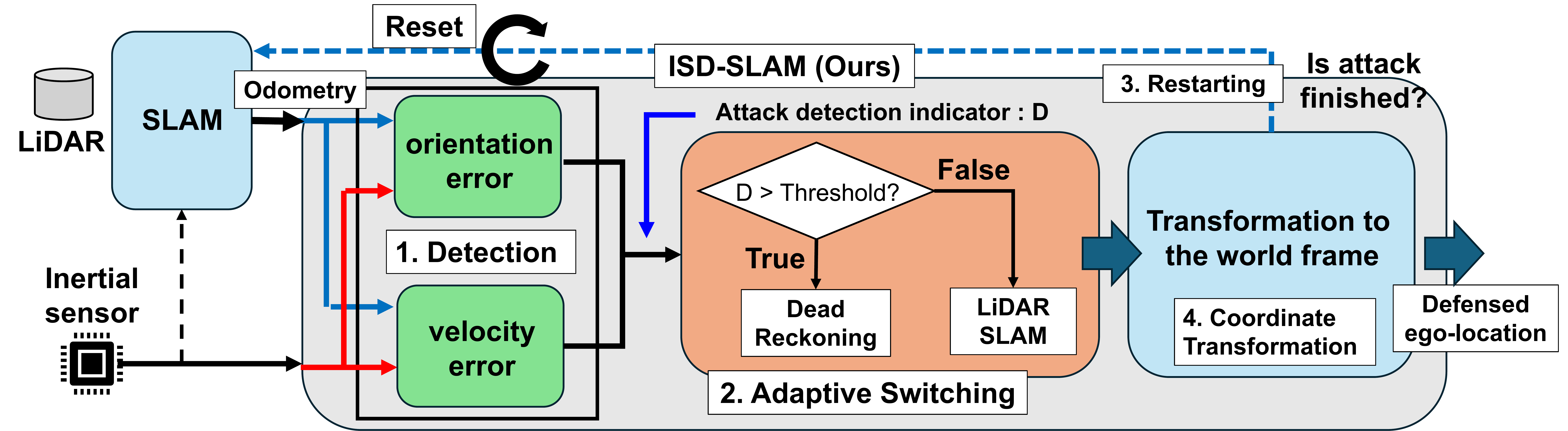}
  \vspace{-0.1in}
  \caption{Overview of ISD-SLAM. The system defends against LiDAR spoofing through four modules: (1) detection of localization errors, (2) switching to inertial navigation, (3) restarting SLAM after the attack, and (4) coordinate transformation to ensure trajectory consistency.}
  \vspace{-0.2in}
  \label{fig_proposed_defense_method}
\end{figure*}

The orientation error $E_{\text{orientation}}$ is defined as the normalized inner product between the estimated and dead-reckoning vectors. 
The speed error detects abrupt drifts or unnatural stops caused by LiDAR spoofing. The speed error, $E_{\text{speed}}$, is derived from the norm of the error between the speed calculated from self-localization and the speed calculated from inertial sensors at the same time. 

We integrate the orientation error and the speed error to obtain an attack detection metric that comprehensively covers the localization error. Since the value ranges of the orientation and speed errors differ, we define the attack detection metric $D$ as shown below. 
\vspace{-0.05in}
\begin{align}
    D = w_{\text{ori}}E_{\text{orientation}}+w_{\text{speed}}E_{\text{speed}}
\end{align}
The metric $D$ represents the magnitude of the  localization error caused by spoofing. By setting a threshold, the detection can be reduced to a binary classification problem.
The weights $w_{\text{ori}}$, $w_{\text{speed}}$ and the threshold are optimized in advance through simulation, and in this study, we performed Bayesian optimization using Optuna~\cite{akiba2019optuna}.

\subsection{Overview of ISD-SLAM}
\label{defence_module_proposal}

Fig.~\ref{fig_proposed_defense_method} shows an overview of the proposed ISD-SLAM.
ISD-SLAM defends the SLAM system from spoofing attacks through the following steps.\\
\noindent\textbf{1. Detection:} An attack is detected when the metric $D$ exceeds a predefined threshold.
\\\noindent\textbf{2. Switching:} Once an attack is detected, the SLAM output is discarded, and the system switches to dead-reckoning based on inertial sensor estimation.
\\\noindent\textbf{3. Restart:} When the attack ends, the SLAM process is restarted to reset the accumulated estimation error.
\\\noindent\textbf{4. Coordinate Transformation:} To maintain consistency with the previous trajectory, the final pose ${^\text{world}}{\bm T}_{\text{current}}$ is estimated using the pose immediately before the restart (${^\text{world}}{\bm T}_{\text{det}}$), the predicted transformation between the attack detection and the SLAM restart (${^\text{det}}{\bm T}_{\text{restart}}$), and the ego-motion estimation after the reset (${^\text{restart}}{\bm T}_{\text{current}}$):
\begin{align}
{^\text{world}}{\bm T}_{\text{current}} = {^\text{world}}{\bm T}_{\text{det}} {^\text{det}}{\bm T}_{\text{restart}} {^\text{restart}}{\bm T}_{\text{current}}
\end{align}
We use the constant velocity model to estimate ${^\text{det}}{\bm T}_{\text{restart}}$. 
By combining attack detection with adaptive sensor switching and trajectory reconciliation, ISD-SLAM suppresses spoofing-induced drift while requiring only standard onboard inertial sensors commonly available on commercial autonomous platforms.
\vspace{-0.1in}

\section{Evaluation}
\label{Evaluation}

\subsection{Conventional static injection vs D-SLAMSpoof}
\label{static_dynamic_spoofing}

This section evaluates whether D-SLAMSpoof can induce localization errors in geometrically diverse environments and compares it against conventional static injection.

\noindent\textbf{Experimental setting (Simulation):}
We simulated placing spoofers at 100 different locations along each of the three courses in the MCD tuhh dataset~\cite{nguyen2024mcd}. D-SLAMSpoof was compared against static cylindrical injection, which serves as the primary real-world-demonstrated baseline among prior LiDAR SLAM spoofing attacks~\cite{nagata_slamspoof_journal}.
The performance was evaluated across three distinct SLAM algorithms: KISS-ICP~\cite{vizzo2023ral}, FAST-LIO2~\cite{fastlio2}, and GLIM~\cite{koide2024glim}.

For the spoofing pattern of static spoofing, a cylindrical wall was injected at a distance of 50 m from the LiDAR, spanning 80$^\circ$ in the horizontal azimuth, as illustrated in Fig.~\ref{header_realworld_experiment}. For D-SLAMSpoof, the minimum injection distance was set to 1.0 m, the maximum injection distance to 50 m, and the injection period to 5 s, while the modifiable azimuthal range was the same as in static spoofing 80$^\circ$. The manipulated point clouds were fed into each SLAM pipeline, and the resulting trajectories were recorded.

\noindent\textbf{Metric:} Attack performance is quantified using Attack Success Rate (ASR) based on Absolute Pose Error (APE)~\cite{zhang2018tutorial}. For each condition, we perform 100 trials; given a threshold $\tau = 3\text{m}$ , a trial is considered successful if $\mathrm{APE}\ge \tau$, and $\mathrm{ASR}_\tau$ is the percentage of successful trials.

\begin{figure*}[tb]
  \centering
  \includegraphics[width=0.98\linewidth]{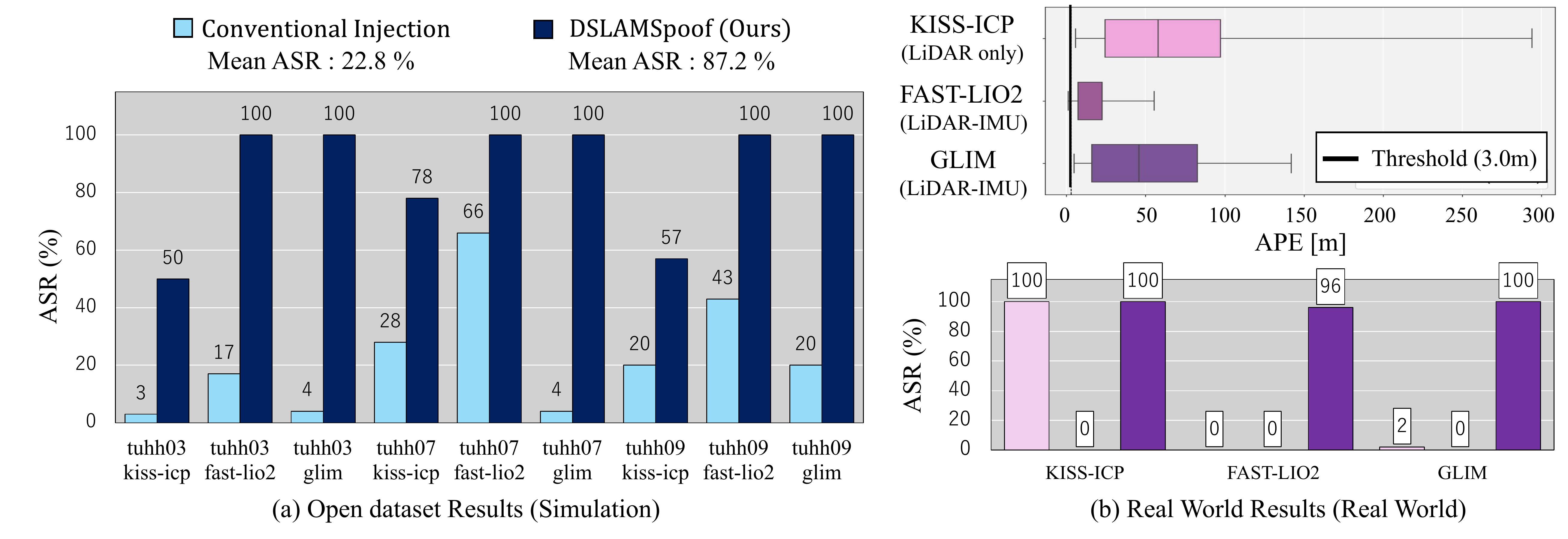}
  \vspace{-0.1in}
  \caption{Comparison between conventional static injection and the proposed D-SLAMSpoof attack. Attack Success Rate (ASR) values are expressed in \%. 
(a) Comparison of ASR on an open dataset (MCD TUHH dataset) in simulation. D-SLAMSpoof achieves an ASR 3.82 times higher than that of the conventional static injection. 
(b) Comparison of ASR in real-world experiments (Sec.~\ref{real_world_experiment}). The ASR is defined as the ratio of trials exceeding the threshold (3.0~m) over 100 trials with randomized seeds. D-SLAMSpoof is the only method that successfully compromises all three SLAM algorithms with high probability.}
\vspace{-0.1in}
  \label{fig_static_vs_dynamic}
\end{figure*}

\noindent \textbf{Results:}
Fig.~\ref{fig_static_vs_dynamic}(a) shows that the baseline static injection achieves an average success rate of 22.8\%. In contrast, the proposed Attack significantly enhances the ASR to an average of 87.2\%, representing a 3.8$\times$ improvement over the baseline. Notably, this increase in attack success rate was consistently observed across all three evaluated SLAM algorithms, demonstrating the broad effectiveness of our proposed method. %

\subsection{Analysis of the Effect of Injection Shapes}
\label{injection_shape_analysis}

This section isolates the contribution of the proposed injection geometries by comparing different injected point cloud shapes under controlled conditions.

\begin{figure*}[tb]
  \centering
  \includegraphics[width=1.0\linewidth]{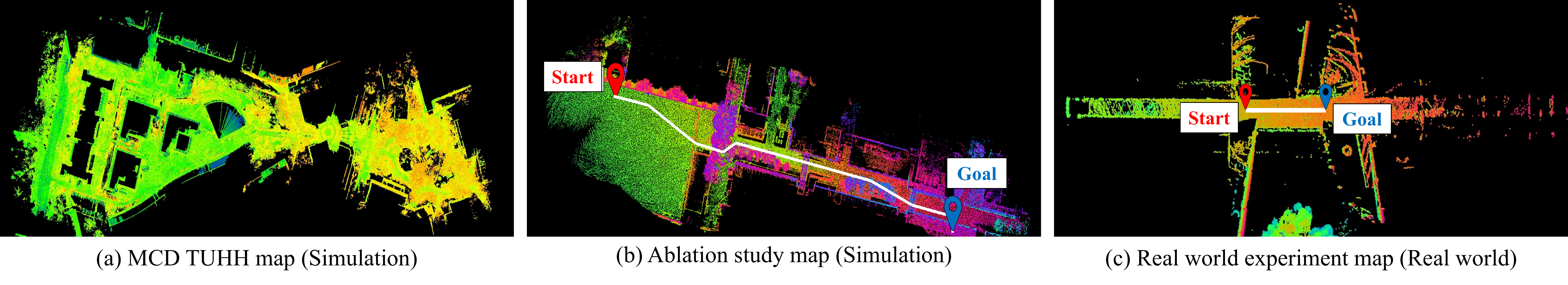}
  \vspace{-0.3in}
  \caption{Experimental environment. (a) MCD TUHH map. The start and goal positions vary depending on the scenarios (tuhh\_03, 07, and 09); the map shown here encompasses the entire area covered by all scenarios.(b)Ablation study map. The starting point is set as the origin $(x=0, y=0)$, and the trajectory proceeds toward the lower-right direction (positive x-axis, negative y-axis). (c) Real-world experiment map. The start position is set as the origin, consistent with (b), and the vehicle travels straight along the positive x-axis. }
  \vspace{-0.2in}
  \label{experimental_environment}
\end{figure*}

\noindent \textbf{Experimental setting (Simulation):}
The experiments were conducted along the trajectory shown in Fig. \ref{experimental_environment}, with spoofers positioned at six locations. To compare the attack capability across different injection shapes, we simulated varying the modifiable azimuthal range from 10$^\circ$ to 80$^\circ$ in increments of 10$^\circ$. The evaluation of the relationship between injected point cloud shape and attack capability was based on self-localization results obtained by FAST-LIO2~\cite{fastlio2}, in the same manner as \S~\ref{static_dynamic_spoofing}.
Injection shapes included the conventional cylindrical wall (Fig.~\ref{header_realworld_experiment}), and our proposed corner-shaped and planar walls. Parameters ($d_{\text{min}}, d_{\text{max}}, t_{\text{cycle}}$) matched those in \S~\ref{static_dynamic_spoofing}.

\noindent\textbf{Metric:}
The evaluation metric is the mean Attack Success Rate across the six locations. For each experimental condition, we ran 50 simulations and used $\mathrm{ASR}_{\tau \geqq 5}$ as the success rate (\%).

\begin{figure}[tb]
  \centering
  \includegraphics[width=0.97\linewidth]{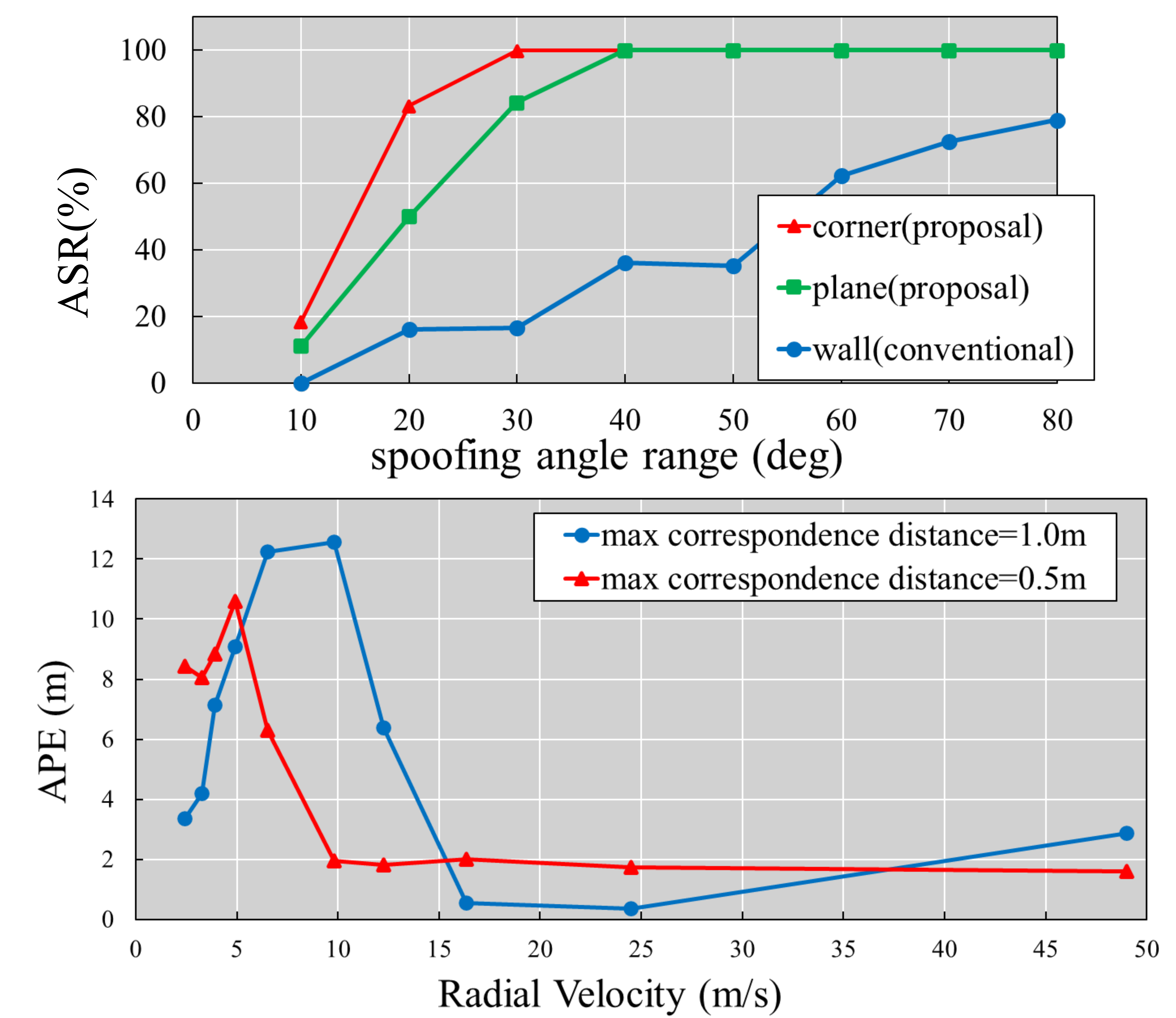}
  \vspace{-0.1in}
  \caption{(Top) Attack success rate (ASR) versus injection shape under limited modifiable azimuth. Proposed shapes maintain high ASR even with a narrow attack range (e.g. $40^{\circ}$), where the conventional method may fail. (Bottom) Attack effectiveness versus radial velocity for different correspondence gates($d_{\text{max}}=1.0$m and $0.5$m) The peak occurs near the gate boundary (9.8 m/s for 1.0 m, 4.9 m/s for 0.5 m), indicating correspondence gating dominates the effect.}
  \vspace{-0.1in}
  \label{result_ablation_study}
\end{figure}

\noindent \textbf{Results:}
Figure~\ref{result_ablation_study} shows that both the proposed corner-shaped and planar injections consistently outperform the conventional cylindrical wall.
For the cylindrical injection, ASR decreases substantially as the modifiable azimuthal range narrows, indicating sensitivity to angular coverage. In contrast, the proposed geometries maintain higher success rates even under restricted azimuthal ranges.
Notably, effective attacks remain achievable with approximately half the angular range required by the cylindrical injection. These results confirm that aligning geometric constraints, rather than merely increasing injected area, is critical for inducing persistent pose errors.

\subsection{Study on Dynamic Spoofing Cycle}
\label{dynamic_spoofing_cycle}
To validate that differences in attack capability are governed by the maximum correspondence distance, as discussed in \S~\ref{attack design policy}, we conducted simulations varying the spoofing cycle $t_{\text{cycle}}$.

\noindent \textbf{Experimental setting (Simulation):}
We conducted simulations on the same course used in \S~\ref{static_dynamic_spoofing}, and selected a scene \#3. The cycle $t_{\text{cycle}}$ was varied from 1 to 20[sec/cycle].
Injection distances were fixed at $r_{\text{min}}=1.0$ m and $r_{\text{max}}=50.0$ m. Unlike previous experiments, here ego-pose estimation was performed using GLIM~\cite{koide2024glim} to analyze the mechanism of dynamic spoofing.

\noindent \textbf{Metric:}
For each experimental condition, the attack effect was evaluated by averaging 25 trials.

\noindent \textbf{Results:}
As shown in Fig.~\ref{result_ablation_study}, the attack effectiveness was strongly influenced by the wall’s moving speed. When the max correspondence distance was 1.0 m, the largest localization error occurred when the radial velocity was 9.8 m/s. When the max correspondence distance was 0.5 m, the largest error occurred at a radial velocity of 4.9 m/s. These results indicate that maximizing the radial displacement of spoofed points within the correspondence distance significantly enhances attack capability.

\subsection{Performance Evaluation of Spoofing Detection System}
\label{spoofing_detection_experiment}

We next evaluated whether the proposed detection method (\S~\ref{propose_inertial_detection}) can identify localization errors caused by spoofing.

\noindent \textbf{Experimental setting (Simulation):}
Spoofers were placed at five locations along the validation course. Training and validation used different courses. Hyperparameters were optimized using optuna~\cite{akiba2019optuna} (300 iterations) under the constraint $w_{\text{ori}} + w_{\text{speed}} = 1$ to enable Precision–Recall analysis.

\begin{figure}[tb]
  \centering
  \includegraphics[width=1.0\linewidth]{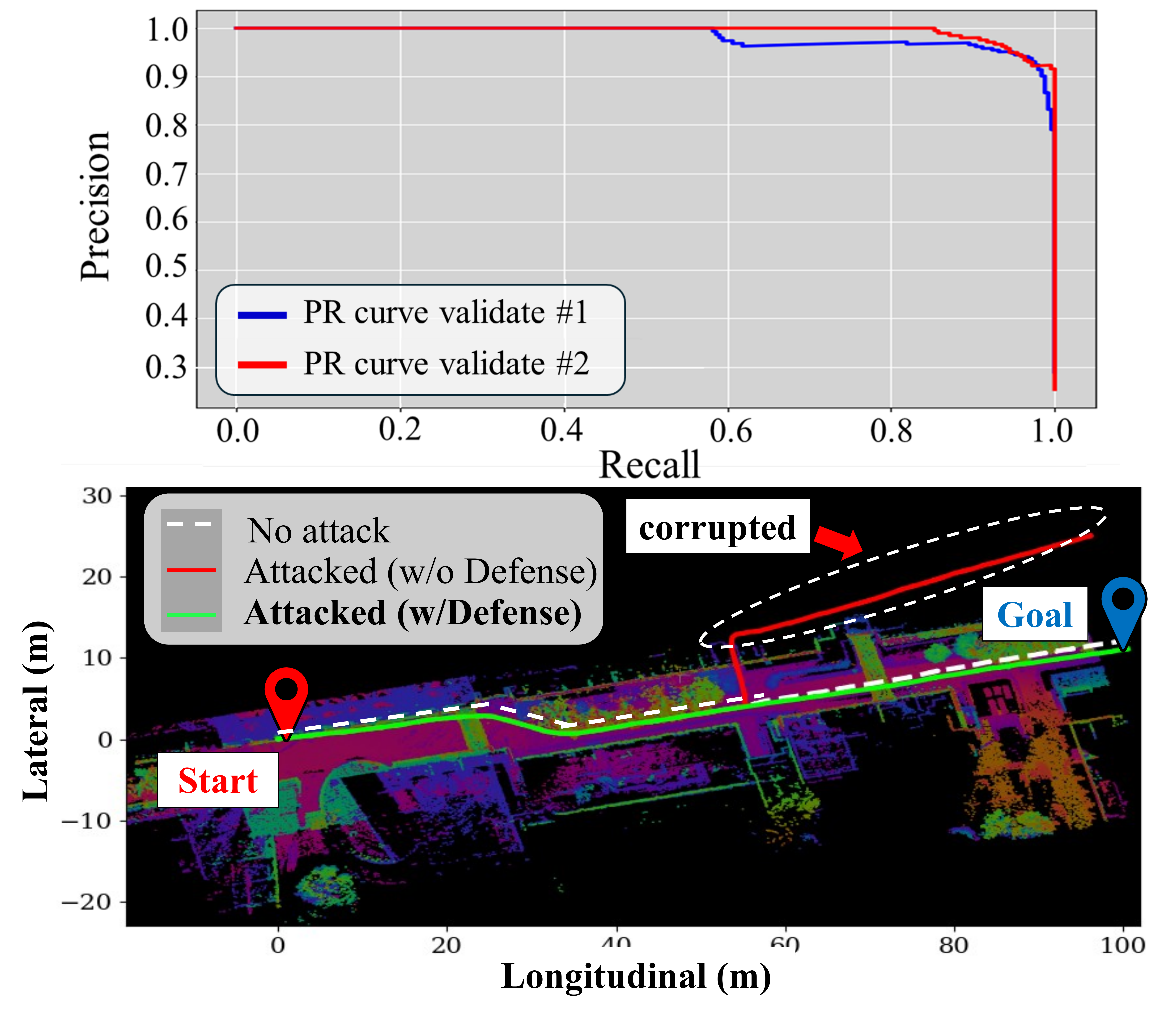}
  \vspace{-0.4in}
  \caption{(Top) Precision–Recall curves on two validation sets (Val\#1/Val\#2), showing consistently high precision and recall across scenarios.
(Bottom) Localization error with and without ISD-SLAM: without defense, spoofing triggers a sharp error increase; with ISD-SLAM, the error growth is suppressed.}
  \label{validate_merge_result}
\end{figure}

\noindent \textbf{Results:}
Fig.~\ref{validate_merge_result} shows that the proposed method demonstrated exceptional detection performance across the validation scenarios through parameter optimization. Specifically, it achieved a Precision of 0.96 and a Recall of 0.97 in both Validation \#1 and \#2, confirming its consistent reliability.

\subsection{Performance Evaluation of ISD-SLAM}
\label{ISD_SLAM_eval}
This section evaluates the effectiveness of the proposed ISD-SLAM (\S~\ref{propose_inertial_detection}) in suppressing localization errors induced by LiDAR spoofing.

\noindent \textbf{Experimental settings (Simulation):}
We simulated attacks on FAST-LIO2~\cite{fastlio2} using D-SLAMSpoof with edge-shaped injections (optimized in \S~\ref{injection_shape_analysis} and \S~\ref{dynamic_spoofing_cycle}). 
Importantly, we reused the same detector parameters (including the weight parameters and the decision threshold) determined in the \S ~\ref{spoofing_detection_experiment} and kept them fixed in this evaluation (i.e., no per-sequence tuning).
Defense effectiveness was measured by comparing trajectories with and without the ISD-SLAM module (Fig.~\ref{fig_proposed_defense_method}).

\noindent \textbf{Results:}
Across two independent validation sequences, ISD-SLAM achieved a Precision of 0.96 and a Recall of 0.97 for attack detection with the fixed parameters.
More importantly, adaptive switching to inertial dead-reckoning effectively suppressed spoofing-induced trajectory drift, preventing the accumulation of large localization errors (Fig.~\ref{validate_merge_result}).

\subsection{Case Study: Real-World Experiment}
\label{real_world_experiment}

Finally, we validated D-SLAMSpoof in a real-world setting.

\begin{figure}[tb]
  \centering
  \includegraphics[width=0.94\linewidth]{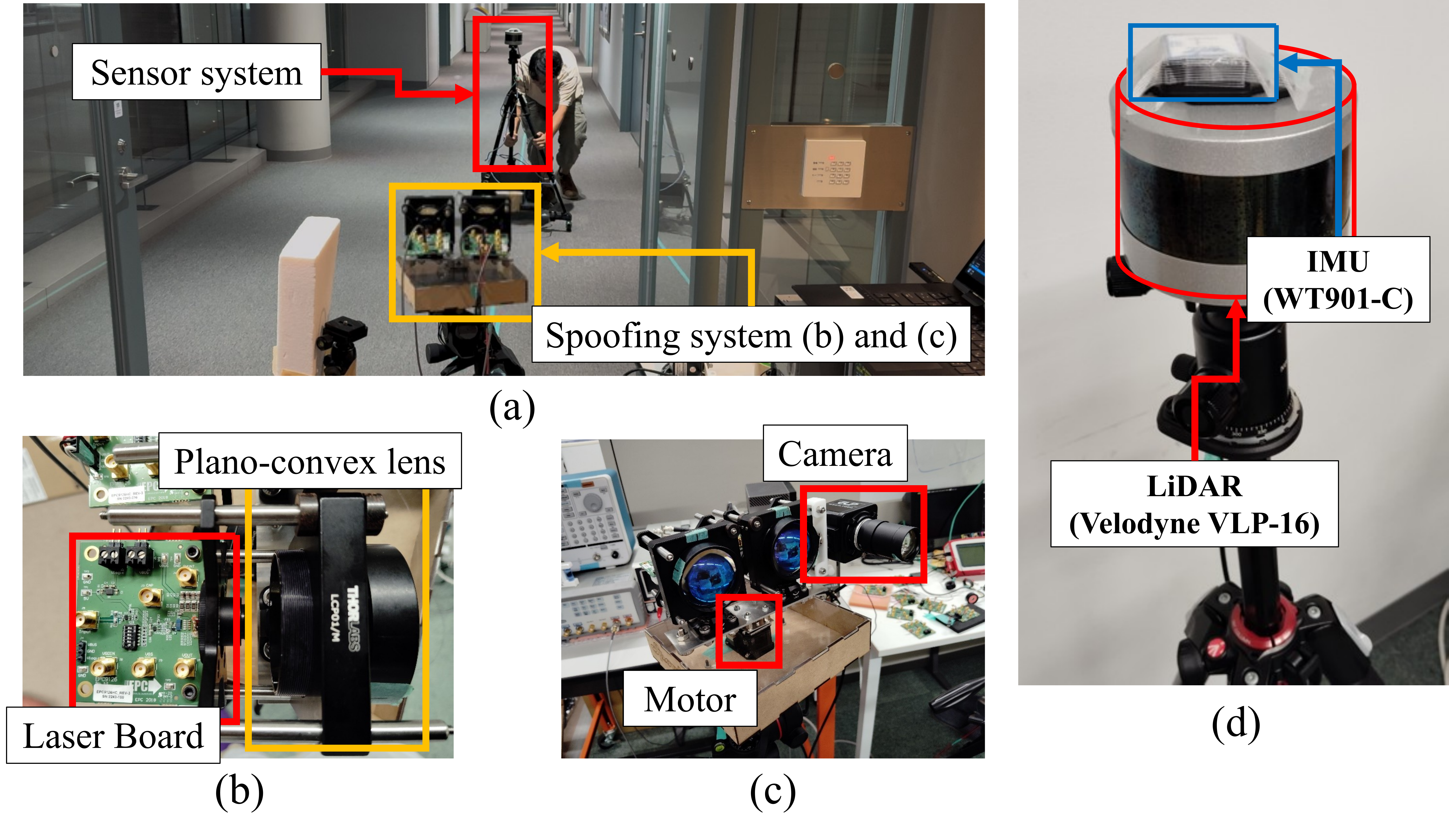}
  \vspace{-0.1in}
  \caption{(a) Real-world experiment setup.The sensor system shown in (d) was operated, and sensor data were recorded when the point cloud was tampered with by the spoofing system. The structure of the spoofing system is shown in (b) and (c). (b) Spoofer configuration. The pulse laser  is collimated by a plano-convex lens. (c) Tracker device~\cite{suzuki2025_from}. A camera detects the target LiDAR, and the motor-driven turntable enables continuous spoofing against moving objects. (d) Sensor system. A mobile-robot-like setup consisting of a VLP-16 LiDAR with an attached IMU (WT-901C).}
  \label{real_world_setting}
\end{figure}

\noindent \textbf{Experimental setting (Real-world):}
We constructed a sensor system using a VLP16 LiDAR and IMU (Fig.~\ref{real_world_setting}). The spoofer design followed~\cite{suzuki2025_from}, enabling dynamic injection of arbitrary point clouds. Experiments were conducted indoors in environments with rich geometrical features. We evaluated attack effectiveness by comparing attacked trajectories with baseline (attack-free) trajectories for the three algorithms used in \S ~\ref{static_dynamic_spoofing}. For each experimental dataset, we performed 100 independent runs (different random seeds/initializations). We report the Attack Success Rate (ASR) using the same definition and threshold as in \S ~\ref{static_dynamic_spoofing}.

\vspace{-0.045in}

\noindent \textbf{Results:} As shown in Fig.~\ref{header_realworld_experiment} and \ref{fig_static_vs_dynamic}(b), D-SLAMSpoof induced severe localization errors in real-world trials; Specifically, the proposed method achieved an attack success rate of nearly 100\% across all three SLAM algorithms: KISS-ICP, FAST-LIO2, and GLIM. This demonstrates that spoofing risks persist even under robust environmental conditions and independent of the spoofer's placement.

\section{DISCUSSIONS}
\label{discussions}

\noindent \textbf{Countermeasures:}
The evaluation results of this study (\S~\ref{Evaluation}) indicate that the introduction of countermeasures against D-SLAMSpoof is essential, regardless of the environment in which the SLAM system is used. In addition to our proposed ISD-SLAM, a promising countermeasure is the use of LiDARs equipped with anti-interference mechanisms (e.g., ranging pulse randomization). Such LiDARs provide effective defense against point cloud injection attacks~\cite{sato2024lidar} and prevent localization errors.\\
\noindent \textbf{Limitations:}
While the proposed attack method D-SLAMSpoof demonstrated high attack capability independent of environmental geometric features, its feasibility is strongly dependent on the victim LiDAR's design. Our method assumes that the LiDAR's scan pattern is known in order to inject arbitrary point cloud shapes, and its effectiveness may be limited against next-generation LiDARs~\cite{yoshioka2022tutorial} equipped with anti-interference mechanisms such as ranging timing randomization or pulse signatures. Future work should consider more general attack realizations that do not rely on LiDAR-specific timing, including relay-style approaches~\cite{petit2015remote}.

Furthermore, as shown in Fig.~\ref{validate_merge_result}(b), we confirmed that our proposed defense method, ISD-SLAM, effectively suppresses the abrupt self-localization errors caused by LiDAR spoofing. One limitation, however, is that determining the optimal detection weight parameters remains challenging when the traversal course is a black box. As a direction for future work, we aim to extend the system to function effectively outside the current threat model, specifically by eliminating the requirement for a predefined trajectory.

\section{CONCLUSION}
\vspace{-0.05in}
We introduced D-SLAMSpoof, a spoofing method that deliberately manipulates scan matching through constraint-forging and oscillating injections. Extensive simulations and real-world trials demonstrated that D-SLAMSpoof  significantly enhances performance in feature-rich environments where conventional methods struggle.

To address this emerging threat, we proposed ISD-SLAM, designed for commercial sensor configurations. By cross-checking LiDAR-based localization with inertial sensing and dynamically switching to inertial navigation, ISD-SLAM successfully detected spoofing-induced errors with over 90\% accuracy in our experiments and mitigated abrupt drifts.

We hope that these contributions provide valuable insights toward the development of secure localization systems and ultimately enhance the safety and reliability of autonomous robotic systems.
\vspace{-0.1in}

\bibliographystyle{IEEEtran}
\bibliography{citation.bib}
\end{document}